# Feature Engineering vs BERT on Twitter Data

Riyaadh GANI [a] Lisa CHALAGUINE [b]
[a] *St John's College, Johannesburg, South Africa*
[b] *University College London, London, United Kingdom*

**Abstract.**
In this paper, we compare the performances of traditional machine learning models using feature engineering and word vectors and the state-of-the-art language model BERT using word embeddings on three datasets. We also consider the time and cost efficiency of feature engineering compared to BERT. From our results we conclude that the use of the BERT model was only worth the time and cost trade-off for one of the three datasets we used for comparison, where the BERT model significantly outperformed any kind of traditional classifier that uses feature vectors, instead of embeddings. Using the BERT model for the other datasets only achieved an increase of 0.03 and 0.05 of accuracy and F1 score respectively, which could be argued makes its use not worth the time and cost of GPU.

## 1. Introduction

The development and publishing of BERT (which stands for Bidirectional Encoder Representation from Transformers) in 2018 [Devlin et al., 2018] had been seen as a revolutionary step in Natural Language Processing (NLP). It achieved state-of-the-art performances on multiple tasks, including sentiment detection, language understanding, question-answering, and text summarisation. BERT being pretrained on an absurd amount of data, and its ability to output contextual word vectors, proved to be consistently better than other language models proposed at the time. At the time of writing, the paper in which BERT was introduced was cited over 46k times.

While BERT has many upsides to it, there are still a few issues surrounding its use. Firstly, due to its size, it is very compute-intensive at inference time, meaning that if you want to use it in production at scale, it can become very costly. And secondly, it requires a substantial amount of data for fine-tuning, which is not always available, and manual labeling is also expensive. This motivates us to use feature engineering. Feature engineering refers to the selection, manipulation, and transformation of raw data into features that can be used to improve the performance of machine learning models. Efficient use of feature engineering in combination with traditional machine learning algorithms could achieve similar results to BERT in certain tasks, making it less time-consuming and cheaper to use in a production environment.

In this work, we compare feature engineering in combination with traditional machine learning methods with BERT for text classification on Twitter tweets using three different open-source datasets. Twitter was chosen because it is nowadays easily one of the most popular microblogging platforms and millions of users express their views publicly on Twitter making it a rich source of information on public opinions and thus, helpful in sentiment analysis and other classification tasks on any topic [Harjule et al., 2020]. The three datasets contain tweets about (1) climate change (2) coronavirus and (3) disasters. We find in the climate change sentiment dataset that the BERT model macro average F1 score was 0.02 higher than our model (0.80 compared to 0.78). For the coronavirus dataset when considering three classes, the BERT model weighted F1 score was 0.1 higher than our model (0.91 compared to 0.81) and when tested on 5 classes, was 0.17 higher than our model (0.79 compared to 0.62). Finally, in the disaster tweet dataset, we compared the accuracy of our model to BERT and found that BERT had an accuracy 4% higher than our model (0.84 compared to 0.80).

The paper is structured as follows: in Section 2 we present related work, Section 3 describes the datasets used, in Section 4 describes our methodology, in Section 5 we present and discuss our results, and finally, we conclude our work and outline future work in Section 6.

## 2. Related Work

Social media platforms such as Twitter are of great interest within the NLP community, specifically due to its enormous amount of variety and rich, real-time conversations which have a uniform length (a maximum of 280 characters). Some examples of previous work include [Kouloumpis et al., 2011] where the authors investigate the utility of linguistic features for detecting the sentiment of Twitter messages; [Kanakaraj and Guddeti, 2015] which analyses the mood of the society on particular news from Twitter posts, and [Hasan et al., 2019] which analyses public views towards a product.

The most prevalent NLP task on Twitter tweets is text classification (e.g. classifying sentiment into positive, neutral, and negative). Commonly used traditional machine learning models for text classification include Multinomial Naive Bayes (MNB) [Kibriya et al., 2004], Logistic regression [Genkin et al., 2007], and Support Vector Machines (SVM) [Joachims, 2002].

Twitter data is usually very noisy and requires certain preprocessing. Preprocessing is effective in removing any extraneous data from tweets that make the model less accurate and inefficient [Krouska et al., 2016]. Preprocessing techniques which have proved effective for tweets include deleting rare words, stemming, and tokenisation. Feature engineering is also effective, which is done by editing the dataset to create features that make machine learning algorithms more accurate. [Scott and Matwin, 1999].

Deep learning methods have brought revolutionary advances in machine learning, including NLP. They can process sequential data in the form of word embeddings, preserving much more meaning of the text than the traditional bag-of-word model. BERT is an NLP model that was designed to pretrain deep bidirectional representations from unlabeled text and, after that, be fine-tuned using labeled text for different NLP tasks [Devlin et al., 2018]. The original English language BERT has two models, a base model and a large one. Both models were pretrained on BooksCorpus (ca 800 million words) and English Wikipedia (ca 2500 million words). Since the models are open source and can be used for free, it allows us to fine-tune them for specific NLP tasks [Devlin et al., 2018].

**Table 1.** Table showing each dataset, its data distribution as well as the number of tweets

| Dataset | Data Distribution | Number of Tweets |
|---|---|---|
| Climate Change Sentiment | Pro: 52% <br> Neutral: 18% <br> Anti: 9% <br> News: 21% | 4123 |
| Coronavirus tweets | Extremely Positive: 16% <br> Positive: 28% <br> Neutral: 19% <br> Negative: 24% <br> Extremely Negative: 13% | 41157 |
| Disaster Tweets | Real Disaster: 43% <br> Not a Real Disaster: 57% | 10876 |

Modern approaches such as the BERT model have two significant drawbacks. The first one is that BERT requires a lot of computational power. Training BERT requires a Graphical Processing Unit (GPU) which can be very expensive [Sun et al., 2019]. The second issue is that fine-tuning BERT requires a significant amount of data which is not always available.

Several papers have compared BERT to traditional machine learning techniques on tweets [González-Carvajal and Garrido-Merchán, 2020, Taneja and Vashishtha, 2022, Benítez-Andrades et al., 2022] and concluded that BERT performs better. However, none of those papers explored feature engineering in combination with traditional machine learning techniques. We argue, that due to its better time and cost efficiency, sometimes feature engineering and preprocessing in combination with a traditional ML model can be a better alternative to using a neural model like BERT.

## 3. Datasets

All the datasets were taken from Kaggle[1] and consisted of tweets from Twitter and their tag towards a specific subject matter. Since we are using supervised learning, the tag is known for the texts. The distributions for each dataset can be seen in Table 1, and an example from each dataset can be seen in Table 2.

### 3.1. *Climate Change Sentiment*

This dataset aggregates tweets pertaining to climate change collected between Apr 27, 2015 and Feb 21, 2018[2]. In total, 43943 tweets were annotated by 3 independent reviewers. Each tweet is labelled as one of the following classes:

- News: the tweet links to factual news about climate change
- Pro: the tweet supports the belief of man-made climate change
- Neutral: the tweet neither supports nor refutes the belief of man-made climate change

---
[1] https://www.kaggle.com/datasets  
[2] https://www.kaggle.com/datasets/edqian/twitter-climate-change-sentiment-dataset

Table 2. Table showing an example text from each dataset and its tag

| Dataset | Text | Tag |
|---|---|---|
| Climate Change Sentiment | @tiniebeany climate change is an interesting hustle as it was global warming but the planet stopped warming for 15 yes while the suv boom | Anti |
| Coronavirus Tweets | Coronavirus Australia: Woolworths to give elderly, disabled dedicated shopping hours amid COVID-19 outbreak https://t.co/bInCA9Vp8P | Positive |
| Disaster Tweets | 13,000 people receive #wildfires evacuation orders in California | Disaster |

- Anti: the tweet does not believe in man-made climate change

It should be noted that although the dataset is officially called "Twitter Climate Change Sentiment Dataset" it is more about *stance* than sentiment, which are two different things. In sentiment analysis, systems determine whether a piece of text is positive, negative, or neutral. Stance detection goes even further and tries to detect whether the author of the text is in favor or against a given target [Hercig et al., 2018].

*3.2. Coronavirus Tweets*

This dataset contains tweets about the Coronavirus in 2020[3]. Manual tagging has been done on the 41157 tweets. Each tweet also contained the date it was tweeted and for 79% of the tweets a location is provided. Each tweet is labelled as one of the following classes: extremely positive, positive, neutral, negative, and extremely negative. The tweets contain the reactions, opinions as well as discussions of people's experiences of the global pandemic in 2020.

*3.3. Disaster Tweets*

The tweets in this dataset are about natural disasters[4]. The dataset is split into a train set of size 7613 and a test set of 3263. The sentiment of the test set cannot be seen since it is part of a kaggle competition. 67% of the tweets in the train set and 66% of the tweets in the test set contain a location, and 99% contain a keyword, although it is not specified how the keywords were chosen. Each tweet is labelled as a real disaster and not real disaster.

In the following section, we will present the methodology we used to preprocess the data and build the classification models.

## 4. Methodology

All datasets were preprocessed and normalised prior to engineering features and applying a machine learning algorithm on them. The Python scikit-learn[5] library was used to implement the classifiers. For each dataset a bag of words model was used because it outperformed the TF-IDF approach. A Multinomial Naive Bayes classifier trained on the raw data and tested on the holdout or test set was used as a baseline to compare to the final results. 10-fold cross validation was used to establish whether a certain preprocessing technique or feature positively impacted the model's performance and for hyperparameter tuning.

---
[3] https://www.kaggle.com/datasets/datatattle/covid-19-nlp-text-classification
[4] https://www.kaggle.com/competitions/nlp-getting-started/data
[5] https://scikit-learn.org/stable/

*4.1. General Preprocessing*

Preprocessing and feature engineering has been a common approach in improving a models performance for machine learning in text classification [Rawat and Khemchandani, 2017, Bao et al., 2014]. Preprocessing often includes stemming and lemmatization [Habash et al., 2009], tokenisation [Straka and Straková, 2017] and removing stopwords [Saif et al., 2014].

Prior to transforming the text into vectors and applying a machine learning algorithm, we considered the following preprocessing methods and experimented with each dataset:

- Tokenisation: Given we were dealing with Twitter data which includes emojis, we used NLTKs[6] Tweet tokeniser to avoid splitting emojis into individual tokens. For example the heart emoji "<3" was kept as an individual token instead of splitting it into "<" and "3".
- Deletion of rare words: Rare words (e.g. typos or user names) are not discriminatory features that a classifier can use to determine the tag and are therefore useless.
- N-gram range: Sometimes it is helpful to not only consider unigrams (single tokens) as features but also bigrams (two tokens), trigrams etc.
- Deletion of stopwords: Stopwords also often do not carry any information and can be removed.
- Stemming and lemmatising: Stemming and helps to create a single feature for the same word with different endings. Stemming never negatively impacted the models performance and hence was applied to all three datasets.

Feature Engineering techniques which are useful for twitter data includes normalisation (replacing numbers, webaddresses etc. with placeholders like "numbr", "emailaddr" etc.), using the tweets location, distinguishing between positive and negative emojis, and many more. Given we were dealing with twitter data where URLs, numbers and other extraneous data were often mentioned we created the following features:

- URLs are replaced with "httpaddr"
- Email addresses are replaced with "emailaddr"
- Numbers were replaced with "numbr"
- Phone numbers were replaced with "phonenumbr"
- Money symbols were replaced with "moneysymb"

---

[6]www.nltk.org

*4.2. Twitter Climate Change Sentiment Dataset*

For this dataset we sampled 20% of the data as a hold-out set and used a Multinomial Naive Bayes classifier to set the baseline on the raw data. The precision, recall and F1 scores for the individual classes are given in Table 3 using the bag-of-words model. Due to the imbalanced data we used the F1 score to evaluate the performance of our model. A TF-IDF approach was also tested using our baseline classifier but the average F1 score was significantly lower (0.51 compared to the 0.64 the bag-of-words model).

**Table 3.** Baseline results for the Climate Change Sentiment dataset for each class as well as the weighted and macro average using the hold-out set.

| Class | Precision | Recall | F1 |
|---|---|---|---|
| Anti | 0.92 | 0.12 | 0.21 |
| Neutral | 0.89 | 0.16 | 0.27 |
| Pro | 0.66 | 0.97 | 0.78 |
| News | 0.85 | 0.66 | 0.74 |
| **Weighted average** | 0.76 | 0.70 | 0.64 |
| **Macro Average** | 0.83 | 0.48 | 0.50 |

Next, we experimented with the (1) threshold of how many times words should appear in the dataset to be included as features (rare word threshold), (2) deletion of stopwords, and (3) whether bigrams should be included as features using 10-fold cross-validation on the training set. The best results were achieved using a minimum threshold of 3 (words that appeared only once or twice in the dataset were deleted), with stopwords kept and bigrams included as additional features, resulting in an average F1 score of 0.71.

We tested several classifiers: multinomial Naive Bayes, logistic regression, support vector machine (SVM), decision trees, random forest and a multi-layer perceptron (MLP). First we used the scikit-learn default hyperparameters to decide which model to proceed with. The results for all the classifiers using the hyperparameteres described above and using 10-fold cross-validation on the training set are shown in Table 4. Logistic regression and MLP performed the best but because the former converged faster we decided to proceed with using a logistic regression classifier.

**Table 4.** F1 scores for tested classifiers on the Climate Change Sentiment dataset.

| Classifier | F1 |
|---|---|
| Naive Bayes | 0.71 |
| Logistic Regression | 0.74 |
| Decision Tree | 0.63 |
| MLP | 0.73 |
| Random Forest | 0.67 |
| SVM | 0.70 |

Next, we preprocessed the data by stemming it and normalising it as described in the previous section (e.g. replacing numbers with placeholders like "numbr"), as well as setting everything to lower case. This resulted in an increase of the F1 score by 0.01. Finally, we tuned the hyperparameters of our model by using a grid search. We considered different types of regularisation (no regularisation, l1, and l2), the strength of the regularisation in a range from 0.1 to 2, the class weights, and the solver. Please consult the scikit-learn documentation for an overview of all hyperparameters[7]. The grid search returned the following best hyperparameters: strong l2 regularisation ($C$ of 0.31) using a newton-cg solver with no class weights.

It should be noted that we tried to enhance performance of our model by dividing the data into news and non-news (non-news being the sentiment categories of pro, neutral and anti) and perform a binary classification first given that news tweets are not *sentimental* and differ from the sentimental tweets. Then we trained a model on the sentimental tweets only and used it on the predicted non-news tweets. However, the chained 2 classifiers did not outperform a single classifier trained on all four categories. The final numbers for precision, recall and F1 scores for the individual classes on the hold-out set using our tuned logistic regression classifier are given in Table 5.

**Table 5.** Results for the Climate Change Sentiment dataset for each class as well as the weighted and macro average using a Logistic Regression classifier on the hold-out set.

| Class | Precision | Recall | F1 |
|---|---|---|---|
| Anti | 0.48 | 0.75 | 0.58 |
| Neutral | 0.50 | 0.61 | 0.55 |
| Pro | 0.87 | 0.79 | 0.83 |
| News | 0.67 | 0.74 | 0.69 |
| **Weighted Average** | 0.79 | 0.77 | 0.78 |
| **Macro Average** | 0.67 | 0.74 | 0.69 |

Finally, we trained a BERT model on the data. BERT does not require any major pre-processing so we trained the model on the raw data. Huggingface's[8] BERT base model (uncased) for sequence classification was used and trained for 3 epochs. This is a nomal BERT model with an added single layer on top for classification. This untrained classification layer is fine-tuned on the given data. The model was saved after each epoch and we evaluated all three models. The results on the hold-out set were all fairly similar with an average F1 score between 0.79 and 0.80. The results for the individual classes using a model trained for 2 epochs can be seen in Table 6.

**Table 6.** Final results for the Climate Change Sentiment dataset for each class as well as the weighted and macro average using BERT on the hold-out set.

| Class | Precision | Recall | F1 |
|---|---|---|---|
| Anti | 0.73 | 0.62 | 0.67 |
| Neutral | 0.50 | 0.72 | 0.59 |
| Pro | 0.88 | 0.82 | 0.85 |
| News | 0.84 | 0.86 | 0.85 |
| **Weighted Average** | 0.81 | 0.79 | 0.80 |
| **Macro Average** | 0.74 | 0.75 | 0.74 |

---

[7]https://scikit-learn.org/stable/modules/generated/sklearn.linear$_m$odel.LogisticRegression.html
[8]https://huggingface.co/

Table 7. Baseline results for all three sentiment classes for the Coronavirus dataset.

| Class | Precision | Recall | F1 | Accuracy |
|---|---|---|---|---|
| Negative | 0.70 | 0.77 | 0.73 | |
| Neutral | 0.77 | 0.12 | 0.20 | |
| Positive | 0.64 | 0.79 | 0.71 | |
| Weighted average | 0.70 | 0.56 | 0.66 | |
| | | | | 0.67 |

## 4.3. Coronavirus Tweets

For this dataset we again sampled 20% of the data as a hold-out set and used a Multinomial Naive Bayes classifier to set the baseline on the raw data. We first set a baseline using all 5 classes, however the extremely negative, extremely positive and neutral classes had very low recall scores (0.09, 0.15, 0.14). We therefore combined extremely negative and negative, and extremely positive and positive. Although it could be argued that neutrality is subjective and not a very important class, [Koppel and Schler, 2006] emphasise the importance of learning neutral sentiment, so we decided to keep it. The baseline results for all three classes can be see in in Table 7.

We preprocessed the dataset, and used a bag-of-word model with a minimum threshold of 20 for word occurrence. We again tested several classifiers where logistic regression performed the best and hence was used for hyperparameter tuning. Results for all tested classifiers can be seen in Table 10

Table 8. F1 and accuracy scores on the Coronavirus Tweets dataset

| Classifier | F1 | Accuracy |
|---|---|---|
| Logistic Regression | 0.80 | 0.80 |
| Naive Bayes | 0.69 | 0.68 |
| MLP | 0.77 | 0.77 |
| SVM | 0.75 | 0.75 |
| Decision Trees | 0.60 | 0.60 |

We tuned the hyperparameters using grid search. We used the same types of regularisation in the previous dataset (0.1 to 2, class weights and the solver). The following were best parameters returned by the gridsearch: C of 0.5, using a sag solver and no classweights. We again trained a BERT model. The final results for our model and BERT can be seen in in Table 9.

Table 9. F1 and accuracy scores on the Coronavirus Tweets dataset using our model and BERT on 3 classes and 5 classes.

| Dataset | Classifier | Logistic Regression | BERT |
|---|---|---|---|
| 3 classes | Weighted F1 | 0.81 | 0.91 |
| 3 classes | Accuracy | 0.81 | 0.91 |
| 5 classes | Weighted F1 | 0.62 | 0.79 |
| 5 classes | Accuracy | 0.62 | 0.79 |

*4.4. Natural Language Processing with Disaster Tweets*

This dataset is part of a Kaggle competition and hence is split into a train and a test set, where the test set does not contain the labels. We used the unlabeled test set to set a baseline on Kaggle using Multinomial Naive Bayes on the raw text data. Kaggle only returns the accuracy, hence we proceeded using accuracy as our performance measure. The baseline was 0.795 accuracy.

Like the other datasets, we experimented with rare word threshold, deletion of stopwords, and if bigrams should be included using 10-fold cross validation on the training set. The best results were achieved using a rare word threshold of 3, with no stopwords and unigrams only, resulting in an average accuracy of 0.80.

We tested several classifiers: Logistic regression, SVM, decision trees, and MLP. We used the default hyperparameters chosen by scikit-learn to decide which model to proceed with. The results for all the classifiers above using 10-fold cross-validation can be seen in Table 10. Although SVM performed slightly better it takes a longer time to converge. Naive Bayes and logistic regression performed the same, however, logistic regression has more hyperparemeters to tune and hence provides a change to increase the performance of the model. We, therefore, proceeded with using logistic regression.

**Table 10.** Table showing the classifier used and its accuracy for the disaster tweets dataset

| Classifier | Accuracy |
|---|---|
| Logistic Regression | 0.805 |
| Naive Bayes | 0.805 |
| MLP | 0.769 |
| SVM | 0.807 |
| Decision Trees | 0.769 |

Preprocessing and cleaning the data resulted in a slight drop in accuracy of 0.01, so we did not apply any preprocessing to this data and used the raw data as input for our model. Using gridsearch again, we tuned the hyperparameters of our model. We considered the different types of regularisation, the strength of the regularisation in a range from 0.1 to 2, the classweights, and the solver. The gridsearch returned the following best hyperparameters: C of 0.31, using a newton-cg solver and no classweights.

We tested our final model on the test.csv dataset and submitted the predictions for the Kaggle competition [9]. The result was an accuracy of 0.801, a jump of 83 places upward on the leaderboard to 280 out of 713[10]

Finally, we trained a BERT model on the data and achieved an accuracy of 0.838 and rank 45 on the leaderboard.

---

[9]https://www.kaggle.com/competitions/nlp-getting-started/data?select=train.csv
[10]Not counting submissions with 0 accuracy and cheaters who obtained 1.0 accuracy by submitting the publicly available targets for the test set.

# 5. Results and Discussion

## 5.1. Twitter Climate Change Sentiment Dataset

As can be seen in Tables 5 and 6, BERT performed slightly better on average, and significantly better on the minority class (anti), as well as the news class which differed from the sentiment classes. Our results are similar to those in [Effrosynidis et al., 2022], the only academic work we could find at the time of writing that used the same dataset. They use a macro average and not a weighted average, hence we will compare their results to our macro averages too: they also used a Naive Bayes classifier as a baseline and achieved a macro averaged F1 score of 0.54, similar to our 0.50 (Table 3. It should be noted, however, that they preprocessed their dataset beforehand by removing URLs, replacing user mentions with the username, removing hashtags and numbers and replacing contractions and elongated words, and including bigrams and trigrams as features. They also used a TF-IDF approach, instead of a bag of words model. When using a Naive Bayes classifier on our preprocessed data, we achieve a macro averaged F1 score of 0.67 - significantly higher. And the macro averaged F1 score for our logistic regression classifier is 0.69 which is 0.04 points higher than the CNNs and LSTM the authors used.

Hence, it can be concluded that appropriate preprocessing and feature engineering results of traditional machine learning models can beat neural methods like CNNs and LSTMs and perform only marginally worse than state-of-the-art models like BERT. Scientists have to weigh the cons and pros when deciding what to opt for. For this particular dataset, if precision is more important than recall, BERT is the preferred model due to its better performance on the minority class. However, BERT requires a GPU and takes much longer to train than logistic regression. Training BERT on this dataset for 2 epochs required 20 minutes, whereas the logistic regression model was trained in less than one minute.

One could argue that neither the results for the logistic regression nor BERT are particularly good, so we inspected what our models got wrong. As with all Twitter data, the data is very noisy and often ambiguous given many tweets are hard to understand out of context. We also do not know what instructions the annotators were given when annotating the data. For example, when should a tweet be labeled as news and when not? From the example below it is not clear why one tweet was labeled as pro, whereas the other was labeled as news.

**Example 1**

- *RT @ClimateNexus: How climate change's effect on agriculture can lead to war https://t.co/ydbrX0v5zJ via @Newsweek https://t.co/2nRVw4YQp0* (true label: pro, predicted label: news)
- News: *RT @climatehawk1: Houston fears climate change will cause catastrophic flooding: 'It's not if, it's when' — @Guardian...* (true label: news, predicted label: pro)

If we had that information, it might be possible to utilise that knowledge to engineer more features or use hand-written rules to boost the model's performance.

## 5.2. Coronavirus tweets NLP - Text Classification

We could only find one paper at the time of writing which used the same dataset [Haynes et al., 2022], however, they work with all 5 classes, while we combined the extreme sentiments with the "non-extreme" sentiments. We decided to do this since it seemed pointless in separating them as the extremely positive and extremely negative categories have an extremely low recall. Although neutral also had a low recall, we still kept it even though neutral is subjective and always difficult to label in any dataset. However, to compare our results with previously obtained results, we also trained our final model on 5 classes and achieved an F1 score of 0.62, whereas the authors only achieved F1 scores of around 0.44 at best.

We identified some tweets with questionable labels presented in Example 2. The first tweet is certainly not positive and tweets 2 and 3 are similar in sentiment but are labeled extremely negative and positive respectively.

**Example 2**

- *I never thought that I would have to explain to a complete stranger that I was coughing due to choking on some spit because I bent down too far instead of having the plague whilst in a supermarket but here we are ?????* (positive)
- *My food stock is not the only one which is empty... PLEASE, don't panic, THERE WILL BE ENOUGH FOOD FOR EVERYONE if you do not take more than you need. Stay calm, stay safe* (extremely negative)
- *Me, ready to go at supermarket during the COVID19 outbreak.Not because I'm paranoid, but because my food stock is litteraly empty. The coronavirus is a serious thing, but please, don't panic. It causes shortage...* (positive)

BERT, however, despite the mislabelled tweets, achieved significantly higher results (an increase of over 0.10 accuracy). When trained on all 5 classes, however, BERT's performance dropped significantly, although not as much as our model's.

It should be noted, however, that recently the BERT model has been trained on coronavirus tweets (CT-BERT) [Jaiman and Bhandari, ]. A model like this can obtain F1 scores which are significantly higher than that of any preprocessing, feature engineering, and traditional machine learning model can probably ever achieve. In [Jaiman and Bhandari, ] the CT-BERT model achieved an F1 score of 0.949.

## 5.3. Natural Language Processing with Disaster Tweets

For this dataset, we found similar results in [Chanda, 2021], the only academic work we could find at the time of writing which used the same dataset. After they had preprocessed their data by removing stopwords and punctuation, and by using a bag-of-words model, their logistic regression classifier gave them an accuracy of 0.79, 0.01 lower than our results. Our bag-of-words model also beats all their models, apart from one which used context-free word embeddings. Their best result was an accuracy of 0.8093 on the Kaggle test dataset using GloVe embeddings and a Bi-LSTM model.

Interestingly, our BERT model performed better than the author's BERT model. Although BERT models achieve an accuracy of around 0.03 higher than traditional models on that dataset, we showed that a traditional bag-of-words model can outperform complex neural networks like LSTM which before BERT, were considered state-of-the-art.

Unfortunately, we are lacking the knowledge of what constitutes a disaster. We found some ambiguous labels in this dataset too as can be seen in Example 3: the first tweet refers to an X-Men movie but is categorised as a disaster. The second tweet is also clearly not a disaster tweet. Hence, one could conclude that with a better-labeled dataset better results could be achieved.

**Example 3**

- *The latest from @BryanSinger reveals Storm is a queen in Apocalypse @RuPaul @AlexShipppp http://t.co/oQw8Jx6rTs* (disaster)
- *Why did I come to work today.. Literally wanna collapse of exhaustion* (disaster)

## 6. Conclusion

In this paper, we explored whether the development of the state-of-the-art BERT model meant the possible end of feature engineering. From our results, we have shown that feature engineering can perform at a similar level to an advanced model such as BERT. This is evident in the climate change dataset where the weighted average of the F1 score was only 0.02 better than feature engineering. It is also evident in the disaster tweet dataset where the BERT model had only outperformed our model from feature engineering by 3.7%. Since BERT requires an expensive GPU, takes longer to train, and requires more data for fine-tuning, this makes feature engineering far more time and cost-efficient than BERT.

In each of our datasets we found the following F1 or accuracy scores for both feature engineering and BERT using the same dataset:

Table 11. Table comparing our model to the BERT Model

| Dataset | Our Model | BERT Model |
| --- | --- | --- |
| Climate Change Sentiment (F1) | 0.78 | 0.80 |
| Coronavirus Tweets (F1) | 0.81 | 0.91 |
| Disaster Tweets (Accuracy) | 0.80 | 0.84 |

From Table 11 we can see that the time and cost efficiency of feature engineering for the Climate Change Sentiment and Disaster Tweet dataset proves to be worth it, however, we see in the Coronavirus tweets dataset that the BERT model outperformed our model by 10%. This significant increase in performance can be said to be worth both the extra time and money invested in using the BERT model.

In this paper, we only considered the standard BERT. A BERT model such as the CT-BERT [Jaiman and Bhandari, ] which was trained on very specific data that applies to that specific dataset performs significantly better to the point where buying a GPU and fine-tuning the model for a long time may be worth the time and cost.

In the future, we hope to see an advanced model such as BERT become more accessible to everyone. In the meanwhile, feature engineering will remain an import part of many NLP tasks, but if the BERT model becomes as time and cost efficient as feature engineering, feature engineering may become obsolete.

# References

[Bao et al., 2014] Bao, Y., Quan, C., Wang, L., and Ren, F. (2014). The role of pre-processing in twitter sentiment analysis. In *International conference on intelligent computing*, pages 615–624. Springer.

[Benítez-Andrades et al., 2022] Benítez-Andrades, J. A., Alija-Pérez, J.-M., Vidal, M.-E., Pastor-Vargas, R., and García-Ordás, M. T. (2022). Traditional machine learning models and bidirectional encoder representations from transformer (bert)–based automatic classification of tweets about eating disorders: Algorithm development and validation study. *JMIR Medical Informatics*, 10(2):e34492.

[Chanda, 2021] Chanda, A. K. (2021). Efficacy of bert embeddings on predicting disaster from twitter data. *arXiv preprint arXiv:2108.10698*.

[Devlin et al., 2018] Devlin, J., Chang, M.-W., Lee, K., and Toutanova, K. (2018). Bert: Pre-training of deep bidirectional transformers for language understanding. *arXiv preprint arXiv:1810.04805*.

[Effrosynidis et al., 2022] Effrosynidis, D., Karasakalidis, A. I., Sylaios, G., and Arampatzis, A. (2022). The climate change twitter dataset. *Expert Systems with Applications*, page 117541.

[Genkin et al., 2007] Genkin, A., Lewis, D. D., and Madigan, D. (2007). Large-scale bayesian logistic regression for text categorization. *technometrics*, 49(3):291–304.

[González-Carvajal and Garrido-Merchán, 2020] González-Carvajal, S. and Garrido-Merchán, E. C. (2020). Comparing bert against traditional machine learning text classification. *arXiv preprint arXiv:2005.13012*.

[Habash et al., 2009] Habash, N., Rambow, O., and Roth, R. (2009). Mada+ tokan: A toolkit for arabic tokenization, diacritization, morphological disambiguation, pos tagging, stemming and lemmatization. In *Proceedings of the 2nd international conference on Arabic language resources and tools (MEDAR), Cairo, Egypt*, volume 41, page 62.

[Harjule et al., 2020] Harjule, P., Gurjar, A., Seth, H., and Thakur, P. (2020). Text classification on twitter data. In *2020 3rd International Conference on Emerging Technologies in Computer Engineering: Machine Learning and Internet of Things (ICETCE)*, pages 160–164. IEEE.

[Hasan et al., 2019] Hasan, M. R., Maliha, M., and Arifuzzaman, M. (2019). Sentiment analysis with nlp on twitter data. In *2019 International Conference on Computer, Communication, Chemical, Materials and Electronic Engineering (IC4ME2)*, pages 1–4. IEEE.

[Haynes et al., 2022] Haynes, C., Palomino, M. A., Stuart, L., Viira, D., Hannon, F., Crossingham, G., and Tantam, K. (2022). Automatic classification of national health service feedback. *Mathematics*, 10(6):983.

[Hercig et al., 2018] Hercig, T., Krejzl, P., and Král, P. (2018). Stance and sentiment in czech. *Computación y Sistemas*, 22(3):787–794.

[Jaiman and Bhandari, ] Jaiman, A. and Bhandari, P. Analysis of covid-twitter-bert.

[Joachims, 2002] Joachims, T. (2002). *Learning to classify text using support vector machines*, volume 668. Springer Science & Business Media.

[Kanakaraj and Guddeti, 2015] Kanakaraj, M. and Guddeti, R. M. R. (2015). Performance analysis of ensemble methods on twitter sentiment analysis using nlp techniques. In *Proceedings of the 2015 IEEE 9th international conference on semantic computing (IEEE ICSC 2015)*, pages 169–170. IEEE.

[Kibriya et al., 2004] Kibriya, A. M., Frank, E., Pfahringer, B., and Holmes, G. (2004). Multinomial naive bayes for text categorization revisited. In *Australasian Joint Conference on Artificial Intelligence*, pages 488–499. Springer.

[Koppel and Schler, 2006] Koppel, M. and Schler, J. (2006). The importance of neutral examples for learning sentiment. *Computational intelligence*, 22(2):100–109.

[Kouloumpis et al., 2011] Kouloumpis, E., Wilson, T., and Moore, J. (2011). Twitter sentiment analysis: The good the bad and the omg! In *Proceedings of the international AAAI conference on web and social media*, volume 5, pages 538–541.

[Krouska et al., 2016] Krouska, A., Troussas, C., and Virvou, M. (2016). The effect of preprocessing techniques on twitter sentiment analysis. In *2016 7th international conference on information, intelligence, systems & applications (IISA)*, pages 1–5. IEEE.

[Rawat and Khemchandani, 2017] Rawat, T. and Khemchandani, V. (2017). Feature engineering (fe) tools and techniques for better classification performance. *International Journal of Innovations in Engineering and Technology (IJIET)*, 8(2).

[Saif et al., 2014] Saif, H., Fernandez, M., He, Y., and Alani, H. (2014). On stopwords, filtering and data sparsity for sentiment analysis of twitter.

[Scott and Matwin, 1999] Scott, S. and Matwin, S. (1999). Feature engineering for text classification. In *ICML*, volume 99, pages 379–388. Citeseer.

[Straka and Straková, 2017] Straka, M. and Straková, J. (2017). Tokenizing, pos tagging, lemmatizing and parsing ud 2.0 with udpipe. In *Proceedings of the CoNLL 2017 Shared Task: Multilingual Parsing from Raw Text to Universal Dependencies*, pages 88–99.

[Sun et al., 2019] Sun, F., Liu, J., Wu, J., Pei, C., Lin, X., Ou, W., and Jiang, P. (2019). Bert4rec: Sequential recommendation with bidirectional encoder representations from transformer. In *Proceedings of the 28th ACM international conference on information and knowledge management*, pages 1441–1450.

[Taneja and Vashishtha, 2022] Taneja, K. and Vashishtha, J. (2022). Comparison of transfer learning and traditional machine learning approach for text classification. In *2022 9th International Conference on Computing for Sustainable Global Development (INDIACom)*, pages 195–200. IEEE.